# DMF-Net: Dual-Branch Multi-Scale Feature Fusion Network for copy forgery identification of anti-counterfeiting QR code

Zhongyuan Guo[1], Hong Zheng[1,*], Changhui You[2], Tianyu Wang[1], Chang Liu[1]

[1] Electronic Information School, Wuhan University, Wuhan, 430072, China

[2] School of Cyber Science and Engineering, Wuhan University, Wuhan 430072, China

* Correspondence: zh@whu.edu.cn

**Abstract:** Anti-counterfeiting QR codes are widely used in people's work and life, especially in product packaging. However, the anti-counterfeiting QR code has the risk of being copied and forged in the circulation process. In reality, copying is usually based on genuine anti-counterfeiting QR codes, but the brands and models of copiers are diverse, and it is extremely difficult to determine which individual copier the forged anti-counterfeiting code come from. In response to the above problems, this paper proposes a method for copy forgery identification of anti-counterfeiting QR code based on deep learning. We first analyze the production principle of anti-counterfeiting QR code, and convert the identification of copy forgery to device category forensics, and then a Dual-Branch Multi-Scale Feature Fusion network is proposed. During the design of the network, we conducted a detailed analysis of the data preprocessing layer, single-branch design, etc., combined with experiments, the specific structure of the dual-branch multi-scale feature fusion network is determined. The experimental results show that the proposed method has achieved a high accuracy of copy forgery identification, which exceeds the current series of methods in the field of image forensics.

## 1. Introduction

Quick Response Code (QR code) is currently the most widely used two-dimensional code [1-2], it uses a certain geometric figure to record data symbol information on a flat black and white figure according to a certain rule, and it has the characteristics of security and flexibility, high information capacity, low printing cost, easy to make, and long-lasting and other advantages. With the rapid economic development and the rise of Internet of Things technology, QR code technology has a relatively complete application condition foundation and related industrial supporting system [3-4]. In the field of product anti-counterfeiting traceability, the QR code traceability technology enters basic information such as the origin, manufacturer, and shelf life of the product into the QR code, and forms a product information database on the cloud server. Consumers can easily complete product identification and comparison, authenticity verification and traceability functions through smartphone scanning [5-7].

However, commodities, especially valuable commodities, often cannot escape the coveting of counterfeiters. Counterfeiters will not only forge the commodity itself, but also forge the anti-counterfeiting QR code on the packaging of the commodity. The most common method of forgery is copy forgery, because the copy forgery method is not only effective, but also low-cost and easy to implement, the copy forgery process is as follows: First obtain the authentic anti-counterfeit QR code on the authentic product, and then copy it with a high-precision copier to obtain the forged copy anti-counterfeiting QR code. In order to effectively identify copy forgery, it is necessary to analyze whether the anti-counterfeiting QR code is a copy. However, there are many models and brands of copier equipment on the market, it is impractical to directly the brand, model and individual of the copier. In this case, the idea we intend to adopt is to determine whether the anti-counterfeit QR code to be tested is copied through the equipment type forensics, if it can be concluded that the anti-counterfeit QR code to be tested is produced by a copier, then it must be forged, because the genuine anti-counterfeiting QR code is directly printed from the digital image generated by the anti-counterfeiting QR code generator.

Traditional inspection methods for forged documents include visible light inspection, infrared inspection, ultraviolet inspection, microscopic inspection, chemical inspection, etc., the above methods need to rely on document inspection equipment and professional inspection personnel, which have the disadvantages of high inspection cost and long inspection time, and some of these methods, such as chemicals used in chemical inspections, can even damage files [8].

With the rapid development of computer technology, people began to use digital processing technology to solve the problem of paper document inspection, it first digitized paper documents into digital images, and then used advanced technologies such as image processing and pattern recognition to analyze and collect evidence on paper documents, which can avoid dependence on document inspection equipment and professional inspectors, improve inspection efficiency, and reduce inspection costs, and realize non-destructive inspection. Researchers have carried out a series of work in the digital non-destructive testing of paper documents:

Some research work is conducted on three types of equipment: laser printers, inkjet printers and copiers. Tchan [9-10] used pre-printed gray-scale image blocks to analyze the differences between different device types, and extracted edge sharpness, surface roughness and image contrast features on the image blocks, but this method is only suitable for feature performance verification, it is necessary to experiment on the monochrome image block, and it is impossible to detect the actual file. Schelze [11] proposed to extract the mean and variance of Discrete Cosine Transform (DCT) coefficients in the frequency domain of scanned images as classification features, which can effectively distinguish device types. This method is mainly used to identify whole-page documents and cannot perform character level detection. Dasari [12] et al. used HSV color space characteristics of color printed characters to identify the three types of equipment, and obtained results comparable to traditional document inspection methods.

Other research work is mainly used to distinguish laser printers and inkjet printers. Lampert [13] analyzed the edge roughness and texture characteristics of the laser printed characters and inkjet printed characters, calculated the linear edge roughness, correlation coefficient, and regional difference by comparing the contour lines of the character image with the down-sampled contour lines, and extracted the texture features according to the gray relationship between character image and filtered character image, finally, support vector machine (SVM) is used as a classifier to obtain character-level detection results. Gebhardt [14] used optical character recognition to recognize the content of characters, and then completed the recognition by extracting the characteristics of the character's vertical edge roughness. Schulze [15] analyzed the difference in roughness between laser printed characters and inkjet printed characters, smoothed the edges of the characters, and extracted the gray distribution characteristics, perimeter change characteristics, and pixel distance characteristics of the characters before and after the smoothing to classify and identify. Umadevi [16] models the selected printed text as mixture of three Gaussian models of text, noise and background, and then uses the Expectation Maximization algorithm to obtain the associated patterns and features of the model.

In summary, the digital forensics of the above text documents mainly obtains digital images through scanner, and then use image processing and pattern recognition technology to process and analyze the digital images. For the anti-counterfeit QR code, its copy forgery identification has the following characteristics:

(1) The anti-counterfeit QR code is usually captured by a smartphone, during the scanning process, due to the relative movement between the anti-counterfeit QR code and the smartphone, and errors in focusing operations, slight blurring is inevitable.
(2) Copy forgery usually occurs on the basis of genuine anti-counterfeiting QR codes, the core problem is to verify whether the anti-counterfeiting QR code to be tested is copied. If it is, it must be forged.

From the above analysis, it can be seen that how to design an efficient copy forgery identification method with a certain anti-blur performance has become the key to solving the problem of copy forgery of anti-counterfeit QR codes.

In recent years, deep learning [17-18] has gradually become a research hotspot and mainstream development direction in the field of artificial intelligence, the research and application of deep learning in the field of digital image forensics has become more and more extensive, such as printer source identification of images and documents [19-22]、camera source identification of images[23-25], image tampering detection [26-28], etc. As a promising research field in machine learning, deep learning can avoid the manual design of feature extractor in traditional machine learning methods, and does not require a lot of expert experience, it automatically extracts the rich features of a large amount of data through a convolutional neural network composed of multiple convolutional layers, so as to achieve better detection results. Therefore, the application of deep learning to the identification of copy forgery of anti-counterfeiting QR code is worth a try and is highly anticipated.

Compared with the genuine anti-counterfeiting code, the image quality of copy forged anti-counterfeiting QR codes will be attenuated, and the image definition will be reduced. Specifically, the lines of the QR code patterns are rougher and the burrs on the edges of the QR code patterns are more abundant. After referring to and summarizing a larger number of image forensics methods based on deep learning, this paper proposes a copy forgery method based on dual-branch multi-scale feature fusion network. The main work and contributions are as follows:

(1) As far as we all know, this is the first time that deep learning is applied to the identification of copy forgery of anti-counterfeiting QR codes. The content of anti-counterfeiting QR code patterns is relatively regular, but our proposed method has nothing to do with the image content and is still applicable to digital images and digital texts.

(2) A Dual-Branch Multi-Scale Feature Fusion Network (DMF-Net) for copy forgery identification of anti-counterfeiting QR code is proposed. The proposed method adopts a dual-branch architecture to perform feature extraction and fusion to enhance feature extraction capabilities, the pre-processing layer is used in the dual-branch CNN to fully enlarge the content difference between the copy forged anti-counterfeiting QR code and the authentic anti-counterfeiting QR code in terms of the roughness of pattern lines and richness of burrs, multi-scale features are extracted by setting different convolution kernel sizes in the preprocessing layers of the two branches. Experiments are carried out on the anti-counterfeiting QR code data set collected on mobile phones. The experimental results show that the method used in this paper is effective, and has a certain degree of anti-blur.

## 2. The difference between genuine anti-counterfeiting QR code and copy forged anti-counterfeiting QR code

In order to facilitate the design of subsequent copy forged identification methods, this section analyzes the difference between the genuine anti-counterfeiting QR code and the copy forged anti-counterfeiting QR code. The flowchart of the production and copy forgery of the anti-counterfeiting QR code is shown in Figure 1.

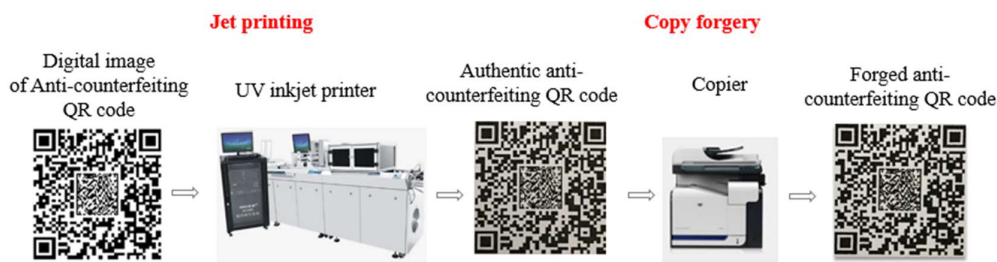

Figure 1. The flowchart of copy forgery

It is worth mentioning that this paper uses Arojet SP-9022 Ultraviolet (UV) inkjet printer when producing authentic anti-counterfeiting codes, as shown in Figure 2.

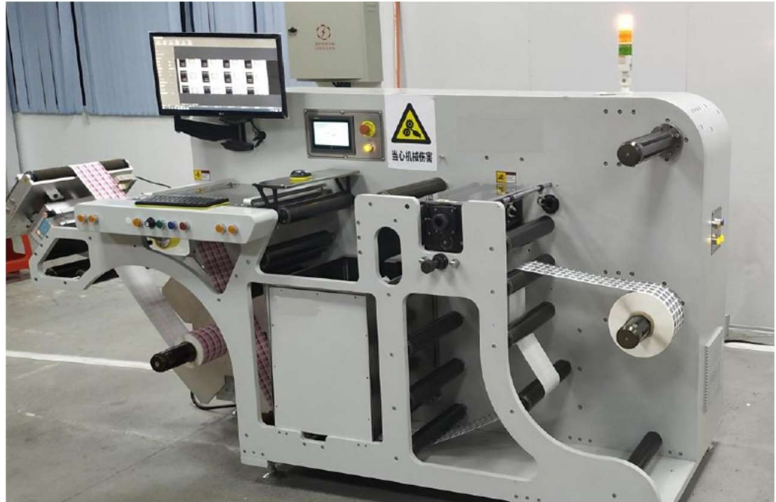

Figure 2.Arojet SP-9022 UV Inkjet Printer

The authentic anti-counterfeiting QR code is printed and produced by Arojet SP-9002 UV digital inkjet system, the system mainly consists of a mobile transport platform for loading paper rolls, an UV inkjet printer and Ricoh industrial piezoelectric nozzles. The width of the nozzle is 54mm, and the longitudinal accuracy of the nozzle is up to 600dpi, the horizontal accuracy is 200dpi-1200dpi, the maximum printing width is 432mm, the ink type used is imported environmental friendly UV ink, and the curing type is LED-UV.

The UV inkjet printer uses hundreds or more piezoelectric crystals to control multiple nozzle holes on the nozzle plate. After processing by the CPU, a series of electrical signals are output to each piezoelectric crystal through the drive board, the piezoelectric crystal generates deformation, the volume of the liquid storage device in the structure will suddenly change, and the ink will be ejected from the nozzle and fall on the moving physical surface to form a dot matrix, thereby forming anti-counterfeiting QR code. After the nozzle discharges ink, the piezoelectric crystal returns to its original state, and new ink enters the nozzle due to the surface tension of the ink. Because of the high density of ink dots per square centimeter, UV inkjet printers can print high-quality anti-counterfeiting QR codes.

Compared with the printing process, the copy forgery process [8] is more complicated. The copier first scans the genuine anti-counterfeiting QR code into an optical analog image, then generates a digital image signal through photoelectric conversion, and finally prints the copy forged anti-counterfeiting QR code. In summary, the copier combines the functions of a scanner and a laser printer, and it will introduce a lot of noise, in addition, copy forged anti-counterfeiting QR codes will inevitably retain the equipment characteristics of UV inkjet printers. As a result, the copy forged anti-counterfeiting QR code contains the inherent characteristics of both UV inkjet printer and copier, which attenuates the quality and definition of the copy forged anti-counterfeiting QR code, the specific performance is that the lines of the anti-counterfeiting QR code are rougher and the image burrs are more abundant.

The comparison between genuine anti-counterfeiting QR code and copy forged anti-counterfeiting QR code is shown in Figure 3.

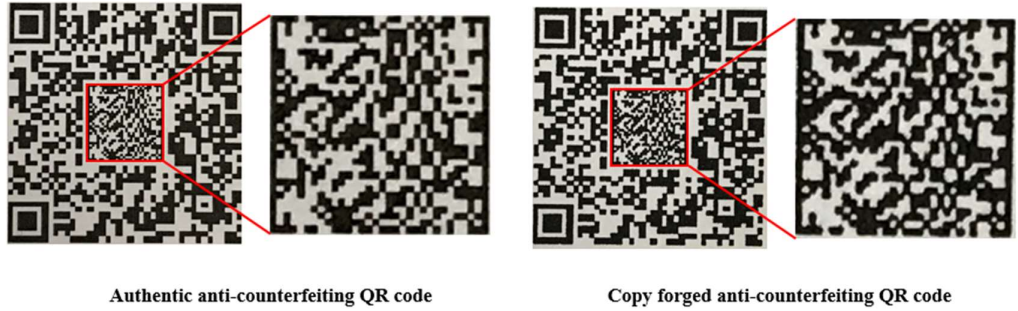

Figure 3. Comparison of genuine anti-counterfeiting QR code and copy forged anti-counterfeiting QR code

## 3. Materials and Methods
### 3.1 Data set production
#### 3.1.1 Copy forged of genuine anti-counterfeiting QR code

We first use the Arojet SP-9022 UV inkjet printer to print 20 anti-counterfeiting codes with different contents as authentic anti-counterfeiting codes, and then use 10 different brands or different models of copiers to copy the 20 authentic anti-counterfeiting codes, and a total of 200 copy forged anti-counterfeiting QR codes are obtained, the model of 10 copiers are shown in Table 1, 20 copy forged anti-counterfeiting QR codes are randomly selected from the 200 copy forged anti-counterfeiting QR codes to ensure that the number of copy forged anti-counterfeiting QR codes is equal to the number of genuine anti-counterfeiting codes.

Table 1. 10 copiers with different brands or models used in this paper

| No. | Brand | Model | Dots Per Inch(dpi) |
|---|---|---|---|
| 0 | HP | M281fdw | 600*600 |
| 1 | HP | M436n | 600*600 |
| 2 | HP | M1136 | 600*600 |
| 3 | HP | P1106 | 600*600 |
| 4 | HP | DJ5078 | 1200*1200 |
| 5 | Canon | MF525dw | 600*600 |
| 6 | Canon | E568 | 4800*1200 |
| 7 | EPSON | L3118 | 5760*1440 |
| 8 | Samsung | C480W | 600*600 |
| 9 | Lenovo | M7268W | 600*600 |

### 3.1.2 Smartphone shooting to obtain data set

Anti-counterfeiting QR codes are often photographed and scanned by mobile phones in actual applications. In order to ensure that the copy forged identification method we designed meets the needs of actual application scenarios, 5 mobile phones are used, their brands and models are shown in Table 2, each mobile phone collected 20 authentic anti-counterfeiting QR Codes and 20 copy forged anti-counterfeiting QR Codes obtained in Section 3.1.1 by shooting, and a total of 200 anti-counterfeiting QR code are obtained, including 100 authentic anti-counterfeiting QR Codes and 100 copy forged anti-counterfeiting QR codes.

Table 2. Five mobile phones with different brands or models

| No. | Brand | Model | Physical pixel |
| --- | --- | --- | --- |
| 0 | Huawei | Honor20 | 48 million |
| 1 | Vivo | iQOO | 12 million |
| 2 | Iphone | 6s | 12 million |
| 3 | Iphone | 6s plus | 12 million |
| 4 | Iphone | 8 | 12 million |

### 3.1.3 Dividing the image into patches

In the field of image forensics based on deep learning, images are generally divided into patches, which are determined by the characteristics of the field of image forensics: the morphological differences between categories are generally extremely small, and these differences exist in a weak form. By dividing the image into patches, the interference of the content can be removed, and the difference between the categories can be enlarged.

The image sized of anti-counterfeiting QR code collected through the smartphone is unified to 512x512. After referring to some research papers in the field of deep learning and image forensics [23, 35, 37], the patch size is determined to be 64x64, and a total of 1,2800 patches of anti-counterfeiting QR codes are obtained, including 6,400 genuine anti-counterfeiting QR code patches and 6,400 copy forged anti-counterfeiting QR code patches.

### 3.2 The design of Dual-branch Multi-scale Feature Fusion Network

In order to highlight the difference in pattern line roughness and burr noise between genuine anti-counterfeiting QR code and copy forged anti-counterfeiting QR code, before the image patches are input to the convolutional neural network (CNN), the preprocessing layer is used, which can suppress the relevant features of the image content and greatly reduce the influence of the image content on the identification of copy forgery. In order to fully extract the multi-scale features that reflect the subtle difference between the genuine anti-counterfeiting QR code and the copy forged anti-counterfeiting QR code, a Dual-branch Multi-scale Feature Fusion Network (DMF-Net) is designed and the size of the convolution kernel of the preprocessing layer in each single branch is different; then the features extracted by the dual-branch CNN are fused; finally the fully connected layer is used to output the classification results. Next, we will introduce in detail the basic unit of the DMF-Net, the structure design of the design of the single-branch CNN, and the specific structure of the DMF-Net

### 3.2.1 Basic unit of DMF-Net

This paper introduces the Bottleneck Residual Block (BRB) as the basic unit of CNN, because the BRB module can effectively improve the depth of CNN while maintaining a small amount of parameters, and has strong feature extraction capabilities. The specific structure of the BRB is shown in Figure 3.

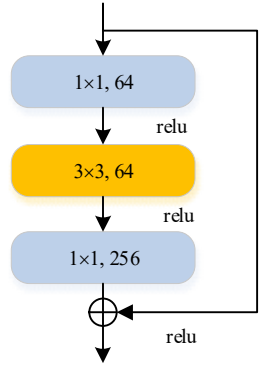

Figure 3. Bottleneck Residual Block

The bottleneck residual block is mainly composed of three convolutions of 1x1, 3x3 and 1x1. The two 1x1 convolutions are used to reduce and increase the number of channels respectively, so that the 3x3 convolution can perform convolution operations with a relatively low-dimensional input, which reduces the amount of parameters while improving computational efficiency; in addition, short connections are completed through simple identify mapping to avoiding introducing more parameters and reducing the computational consumption.

### 3.2.2 Design of the Single-Branch CNN in DMF-Net

We try to ensure that the Single-Branch CNN achieves the best identification effect by superimposing the BRB. Assuming that the size of the input image is $(H_{in}, W_{in})$, the size of the output feature map after the convolution operation is $(H_{out}, W_{out})$, the size of the convolution kernel is $k*k$, the stride is $s$, and the pad is $k$, the calculation formula of the feature map is as follows:

$$H_{out} = (H_{in} + 2p - k)/s + 1 \qquad (1)$$

$$W_{out} = (W_{in} + 2p - k)/s + 1 \qquad (2)$$

Since the size of the input image patch is 64x64, according the above formulas, due to the limitation of the size of the input image patch, the number of CNN's layers should not be too deep, so the number of BRB is set to 1, 2 and 3 respectively, the single-branch CNN's structure with the best performance is determined through experiments, and the final determined CNN's structure is shown in Figure 4.

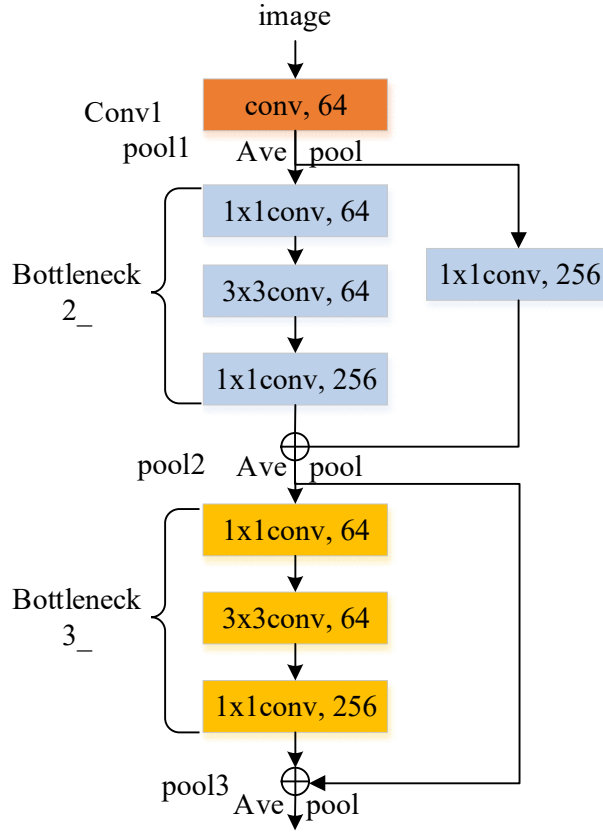

Figure 4. The Single-Branch CNN's structure

The main feature extraction part of the single-branch convolutional neural network in Figure 4 is mainly composed of Conv1, pool1, Bottleneck1_, pool2, Bottleneck2_, pool3 and the fully connected layer. The size of the convolution kernel in the convolutional layer Conv1 is 5x5, the stride is set to 1, and the pad is set to 2, the number of feature map is 64, the ReLU activation layer and the Batch Normalization layer are connected after the convolutional layer, the ReLU activation layer can increase the non-linear feature extraction ability, the formula is as follows:

$$f(x) = max(0, x) \qquad (3)$$

The Batch Normalization layer alleviates the problem of gradient dispersion in the deep CNN to a certain extent, accelerates the convergence speed of the CNN, and makes the CNN's training easier and more stable.

The BRB (Bottleneck1_, Bottleneck2_) is mainly composed of three convolutional layers of 1x1, 3x3, and 1x1, the number of feature maps corresponding to the above three convolutional layers are 64, 64 and 256 respectively. The first 1x1 convolutional layer's stride is set to 1, the pad is set to 0; the stride of the 3x3 convolutional layer is set to 1, and the pad is set to 1; the stride of the last 1x1 convolutional layer is set to 1, and the pad is set to 0; each convolutional layer of the BRB also uses the ReLU activation layer and the Batch Normalization layer.

The pooling layer (Pool1, Pool2) are the average pooling layers, the convolution kernel size of the average pooling layer is set to 5x5, and the stride is set to 2, and the pad is set to 0; the pooling layer Pool2 after Bottleneck2_ is a global average pooling layer, and its convolution kernel size to 6, the stride is set to 1, the pad is set to 0.

### 3.2.2 The Dual-Branch Feature Fusion Network's structure

Since the genuine anti-copy QR code and the forged QR code are the same in content, in order to further highlight the subtle differences between the anti-copy QR codes and the forged QR codes, we draw on the ideas in researches such as median filter forensics [41], camera source forensics [23], and use the preprocessing layer to suppress the interference of image edges and textures on the identification of forgery, and expose the traces left in the process of forgery.

This paper proposes a Dual-branch Multi-scale Feature Fusion Network (DMF-Net) for forgery identification, the proposed network is mainly composed of two single-branch CNNs, a preprocessing layer is added before each single branch, the size of the convolution kernel of the preprocessing layer is specifically determined through experiments, and then the image patches processed by the two preprocessing layers are sent to their respective single-branch CNNs for further feature extraction. The structures of the two single-branch CNNs are the same, the features extracted by the two branches are sent to the fully connected layer for classification after feature fusion, the main categories include genuine anti-copy QR codes and forged anti-counterfeiting QR codes, so the output category of the fully connected layer is set to 2. We will introduce in detail the basic unit of the DMF-Net and the design of single branch structure.

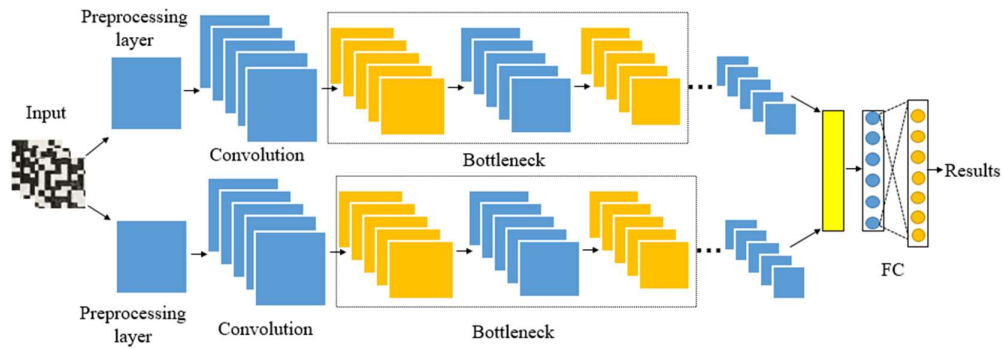

Figure 5. The framework of dual-branch multi-scale feature fusion network

## 4. Experimental results and discussion
### 4.1 Experimental setup

The computer used in the experiment is HP OMEN 17-w119TX, the CPU model is i7-7700HQ (16GB, 2.8GHz), the GPU model is Nvidia Geforce GTX1070 (8GB), the deep learning framework used is Caffe [42], the CUDA version is 9.0 and the version of CUDNN is 7.0.5, it is worth mentioning that we used Digits [43], a web version of the deep learning training tool developed by Nvidia, which can graphically operate and visualize the deep learning model on the web.

### 4.2 The effect of the number of BRBs on the performance of a single-branch convolutional neural network

This paper uses experiments to determine the number of BRBs, and then determines the specific structure of the Single-Branch CNN. Since the sizes of the input image patch is 64x64, its size is small, it can be known from the feature map calculation formula that it is not appropriate to use a CNN with a large number of convolutional layers. Based on the above considerations, the number of BRB is set to 1, 2 and 3 respectively, and then the anti-counterfeiting QR code data set is used for experiments. The hyper-parameters is set as follows: the solver type is set to Stochastic Gradient Descent, the base learning rate is set to 0.01, policy is set to Step Down, step size is set to 33%, Gamma is set to 0.1, the training period is set to 30, and the number of iterations is set to 7200. The influence of the number of BRBs on the identification accuracy is shown in Table 1.

**Table 1. The influence of the number of BRBs on the identification accuracy**

| The number of BRBs | Accuracy (%) |
|---|---|
| 1 | 97.85% |
| 2 | **99.34%** |
| 3 | 98.20% |

From the experimental results, it can be concluded that when the number of BRB is 1, the identification accuracy is 97.85%; when the number of BRB is 2, the identification accuracy is 99.34%, and when the number of BRB is 3, the identification is 98.20%. Therefore, the number of BRBs of the final single-branch CNN is set to 2.

### 4.3 The impact of the specific design of the preprocessing layer on the dual-branch multi-scale feature fusion network

After determining the structure of the single-branch CNN, the next step is to determine the specific structure of the dual-branch multi-scale feature fusion network, especially to analyze how the pre-processing layers of the two branches can be set to achieve the best identification effect.

From the above analysis, we can see that the larger the size of the convolution kernel of the convolutional layer, the smaller the feature map output after the convolution operation. After referring to the relevant literature in the field of image forensics, convolution kernels of 3x3 and 5x5 are selected, and the experimental results of the two preprocessing layers under different settings shown in Table 2.

Table 2. The influence of the setting of the preprocessing layer on the identification accuracy

| The settings of pre-processing layer | Identification accuracy |
|---|---|
| No preprocessing layers | 99.18% |
| One 3x3 preprocessing layer | 99.30% |
| One 5x5 preprocessing layer | 99.26% |
| Two 3x3 preprocessing layers | 99.34% |
| One 3x3 preprocessing layers, One 5x5 preprocessing layers | **99.77%** |
| Two 5x5 preprocessing layers | 99.53% |

It can be seen from the experimental results that when the preprocessing layer is not used, the identification accuracy is 99.18%; when one branch uses the 3x3 preprocessing layer, and the other branch does not use the preprocessing layer, the identification accuracy is 99.30%, when one branch uses a 5x5 preprocessing layer, and the other branch does not use the preprocessing layer, the accuracy is 99.26%; when two branches all use the preprocessing layers with 3x3 convolution kernel size, the identification accuracy is 99.34%; when the convolution kernel sizes of the two pre-processing layers of the dual-branch CNN are 3x3 and 5x5 respectively, the identification is 99.77%; when the size of the convolution kernel sizes of the two preprocessing layers of the two branches are 5x5, the accuracy is 99.53%.The following conclusions can be drawn:
(1) Using the preprocessing layer can improve the identification accuracy.
(2) By using the preprocessing layers with different convolution kernel sizes in the dual-branch structure, the expression of feature scales can be made richer, so as to obtain a better identification effect.

Therefore, we finally determined that the sizes of the convolution kernel of the preprocessing layers in the two branches are 3x3 and 5x5, respectively.

**4.4 Comparison of the proposed DMF-Net and other methods**

After determining the specific structure of the dual-branch multi-scale feature fusion network, other similar algorithms are used to compare with the proposed DMF-Net to verify the superiority of the proposed method.

AlexNet [44], which is currently widely used in the field of digital image forensics, is used as one of the comparison methods. In addition, Anshlesh Sharma [45] proposed a CNN for the authenticity of luxury goods based on AlexNet, the proposed CNN reduces the sizes of the convolution kernel and the stride in the first convolutional layer, and also greatly reduces the batch size, which can improve the ability to extract fine-grained features, due to the modified AlexNet is similar to the proposed DMF-Net in application scenarios, so it is also used as one of the comparison methods. The above methods are tested on the anti-counterfeiting QR code data set, and the results of confusion matrix and identification accuracy are shown below.

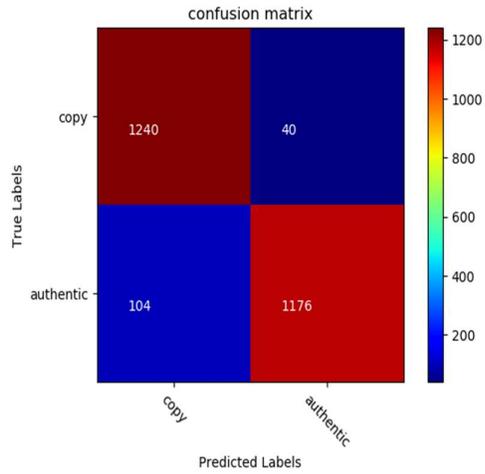

(a) AlexNet's confusion matrix

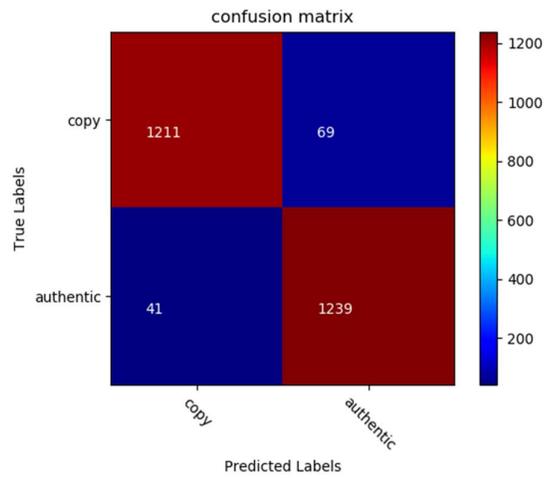

(b) Ashlesh Sharma's confusion matrix

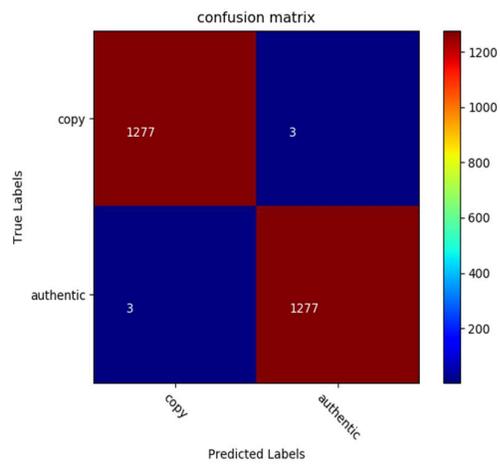

(c) The proposed DMF-Net

Figure 6. Confusion matrix comparison of several methods

In the field of machine learning, each column in the confusion matrix represents the predicted category, and each row represents the actual category of the data, all correct prediction results are on the diagonal, and the results outside the diagonal are the prediction errors. From the diagonal of the confusion matrix of the three methods, it can be directly draw that the DMF-Net proposed in this paper has the largest number of correct identifications, followed by Ashlesh Sharma's method, and AlexNet has the most identification errors.

The identification accuracy of the three methods is shown in Table 3.

Table 3. Comparison of identification accuracy of several methods

| CNNs | Identification accuracy |
|---|---|
| AlexNet | 94.38% |
| Ashlesh Sharma's[45] | 95.7% |
| DMF-Net | **99.77%** |

The identification accuracy of the three methods are shown in Table 3, it can be seen from the Table 3 that the identification accuracy of AlexNet is 94.38%, the identification accuracy of Ashlesh Sharma's method is 95.7%, and the identification accuracy of the proposed DMF-Net reaches 99.77% . The identification accuracy of DMF-Net proposed in this paper is 5.29% higher than that of AlexNet and 4.07% higher than that of Ashlesh Sharma's method, which verifies the effectiveness and superiority of DMF-Net.

5. Conclusion

This paper proposed a copy forgery identification method of anti-counterfeiting QR code based on dual-branch multi-scale feature fusion network. The difference in pattern line roughness and burr noise between the genuine anti-counterfeiting QR code and the copy forged anti-counterfeiting QR code is analyzed, which provided the basis for the design of the CNN's specific scheme, then the structure of single branch CNN is determined through the stacking of BRBs, the dual- branch structure is combined with different settings of the convolution kernel size in the preprocessing layer to perform multi-scale feature fusion, thereby improving the accuracy of copy forgery identification; finally, a series of experimental analysis is done to verify the effect of the DMF-Net proposed in this paper.